\documentclass{article}
\usepackage{PRIMEarxiv}
\usepackage[utf8]{inputenc} 
\usepackage[T1]{fontenc}    
\usepackage{url}            
\usepackage{booktabs}       
\usepackage{amsfonts}       
\usepackage{nicefrac}       
\usepackage{microtype}      
\usepackage{lipsum}
\usepackage{fancyhdr}       
\usepackage{graphicx}       
\usepackage{listings}
\usepackage{mdframed}
\usepackage[round]{natbib}
\usepackage{comment}
\usepackage{tabularx}
\usepackage{caption}
\usepackage{float}
\usepackage{xcolor}
\usepackage{enumerate}

\usepackage{bm}
\usepackage{tikz} 
\usepackage{pifont} 
\usepackage{amssymb}
\usepackage{amsmath,upgreek}
\usepackage{titlesec}
\usepackage{subcaption}
\usepackage[title]{appendix}
\usepackage{stfloats}
\usepackage{multicol,subeqnarray}
\usepackage[most]{tcolorbox}
\definecolor{block-gray}{gray}{0.85}

\definecolor{xlinkcolor}{cmyk}{1,0.6,0,0}
\usepackage[bookmarks=false,         
     pdfnewwindow=true,      
     colorlinks=true,    
     linkcolor=xlinkcolor,     
     citecolor=xlinkcolor,     
     filecolor=xlinkcolor,  
     urlcolor=xlinkcolor,      
final=true
]{hyperref}
\graphicspath{{img/}}     
\usepackage{booktabs} 
\usepackage{array} 
\newfloat{listing}{htbp}{loa}  
\floatname{listing}{Listing} 

\title{A Study of Optimizations for Fine-tuning Large Language Models}

\author{
   \\
  \textbf{Microsoft}\\
Arjun Singh\thanks{Work done while the author was at Microsoft.}\hspace{1mm}\thanks{These authors contributed equally to this work.}\hspace{0.1cm},
Nikhil Pandey\footnotemark[2]\hspace{0.1cm},
Anup Shirgaonkar,
Pavan Manoj,
Vijay Aski
}

\begin{document}
\maketitle

\begin{abstract}
Fine-tuning large language models is a popular choice among users trying to adapt them for specific applications. However, fine-tuning these models is a demanding task because the user has to examine several factors, such as resource budget, runtime, model size and context length among others. A specific challenge is that fine-tuning is memory intensive, imposing constraints on the required hardware memory and context length of training data that can be handled. In this work, we share a detailed study on a variety of fine-tuning optimizations across different fine-tuning scenarios. In particular, we assess  Gradient Checkpointing, Low-Rank Adaptation, DeepSpeed's Zero Redundancy Optimizer and FlashAttention. With a focus on memory and runtime, we examine the impact of different optimization combinations on GPU memory usage and execution runtime during fine-tuning phase. We provide our recommendation on the best default optimization for balancing memory and runtime across diverse model sizes. We share effective strategies for fine-tuning very large models with tens or hundreds of billions of parameters and enabling large context lengths during fine-tuning. Furthermore, we propose the appropriate optimization mixtures for fine-tuning under GPU resource limitations.
\end{abstract}


\section{Introduction}

Transfer learning offers a very effective way to specialize large language models (LLMs) to a particular task or domain. In fine-tuning, the pretrained LLMs are further trained, usually on a task specific dataset, allowing them to adapt their knowledge to the specific task. Although a powerful technique, it is challenging to fine-tune LLMs with several billions of trainable parameters due to their large memory requirement. For example, it requires approximately 24 GB of high bandwidth memory (HBM) per GPU for fine-tuning a 1 billion parameter model in full floating-point precision (32-bit) \citep{fregly2023generative}. Thus, it is common to run into out-of-memory failures during LLM fine-tuning in the absence of any memory optimizations.

In the recent past, several key techniques have been proposed to optimize GPU memory usage. However, they can be complex to understand for many users, and their usage often requires experimentation to find the right combination to avoid out of memory errors. This would necessitate both human resources (to study and run these optimizations), as well as compute resources (large number of GPU hours) to select the right optimizations for a given task. To facilitate the ease of fine-tuning for users, platforms such as \href{https://ml.azure.com/}{Azure Machine Learning} offer state-of-the-art techniques for optimizing GPU memory usage, enabling LLM fine-tuning even with a handful of GPUs. In this paper, we share details about popular optimization techniques for fine-tuning LLMs.  We provide a deep dive into memory and run-time tradeoffs, providing guidance for picking the best optimization configurations, and share our experimental findings which can be used to calibrate optimization defaults. Our goal for this paper is described below:
\begin{itemize}
    \item Demonstrate with empirical results which optimizations work well in achieving memory-efficient fine-tuning while getting the best training time. 
    \item Provide user guidance on selecting fine-tuning optimizations in various settings. The settings we consider include fine-tuning very large models, fine-tuning with large context lengths of training data, and fine-tuning under GPU constraints (SKU or number of GPUs).
\end{itemize}

The remainder of this work is organized as follows. We review the existing fine-tuning optimization literature in section \ref{rw}. Section \ref{oot} presents an overview of commonly used fine-tuning optimizations: Gradient Checkpointing, Low-Rank Adaptation, DeepSpeed's Zero Redundancy Optimizer and FlashAttention. We then describe in Section \ref{tagmr} the theory behind GPU memory usage and compare theoretical values with our empirical results. In Section \ref{exps}, we discuss our experimental results and analyze the impact of using fine-tuning optimizations in conjunction with each other. Finally, we share the key conclusions and directions for future work in Section \ref{cfw}.

\section{Related Work}\label{rw}
Fine-tuning large language models is a computational and memory intensive process. Several optimizations proposed in existing literature either reduce the memory consumption, or the runtime, or accomplish both during fine-tuning. Tensor and pipeline parallelism are types of model parallelism techniques which enable large models to fit in memory for training. In tensor parallelism \citep{shoeybi2020megatronlm}, the model is split tensor-wise among GPUs. Conversely, pipeline parallelism \citep{huang2019gpipe} distributes the model layer-wise across GPUs. DeepSpeed's Zero Redundancy Optimizer (ZeRO) \citep{rajbhandari2020zero} integrates data and model parallelism, by partitioning model states across data-parallel processes, which are reconstructed when required during training. Fully Sharded Data Parallel (FSDP) \citep{FairScale2021} is similar to ZeRO and supports model parameter sharding to train large models.

Contrary to parallelism strategies, Parameter-Efficient Fine-Tuning (PEFT) methods \citep{Ding2023, han2024parameterefficient} modify a small proportion of model parameters, while keeping the others frozen during training. This reduction in total trainable parameters subsequently lowers the memory footprint while still achieving comparable performance to full fine-tuning. Low-Rank Adaptation \citep{hu2021lora} is a widely used PEFT technique, where low rank tensor formulation of model parameters are updated in the training process. Prefix tuning \citep{li2021prefixtuning} and Prompt tuning \citep{lester2021power} are examples of PEFT method where trainable embeddings are injected at the model or input level, while the entire model is frozen. PEFT methods can easily integrate with parallelism techniques such as ZeRO, enhancing the gains in terms of memory and runtime reduction during fine-tuning.

Certain optimizations are designed to address specific memory bottlenecks during fine-tuning. Gradient computation during backward pass can consume significant GPU memory for large models because of the need to maintain large number of activations in memory. Gradient Checkpointing \citep{GC2019} saves only few activations from forward pass, while recomputing most of them during backward pass, to reduce the memory usage at the cost of increased runtime. Attention calculations can form memory bottlenecks when context length is very large. FlashAttention \citep{dao2022flashattention, dao2023flashattention2} methods focus on optimizing the memory consumed during attention calculation operations. Both Gradient Checkpointing and FlashAttention can be combined with PEFT and ZeRO.

Given a wide-variety of optimization methods and diverse use-cases such as fine-tuning very large (multi-billion parameter) models, fine-tuning on lower resource GPUs such as ND40 \citep{nd40rsv2}, or fine-tuning across large context lengths, it becomes important to select the appropriate set of optimizations for fine-tuning. Not all optimizations, even though compatible with each other, may be necessary as they often trade between memory, runtime and accuracy. Practical guidelines for fine-tuning LLMs for enterprise \citep{j2024fine} were generated without leveraging data and model parallelism frameworks like DeepSpeed and do not provide theoretical basis for understanding GPU memory consumption. Despite the differences, those guidelines examine QLoRA \citep{dettmers2023qlora} and can be considered complementary to our study. To the best of our knowledge, it is rare to find studies which offer fine-tuning guidance tailored to specific but common use-cases, and which explore the interaction of state-of-the-art optimizations.

\section{Overview of Optimization Techniques}\label{oot}

We present a brief overview of four optimization techniques that can be used to reduce memory bottlenecks during fine-tuning. All of these optimizations can be leveraged by the users for their fine-tuning jobs on platforms like Azure Machine Learning and Hugging Face.

\subsection{Gradient Checkpointing}

Gradient Checkpointing (GC) \citep{chen2016training} makes judicious use of GPU memory by not preserving all the activations computed during the forward pass (FP) of a Deep Neural Net (DNN). It instead recalculates many activations during the backward pass which helps conserve GPU memory. The most memory efficient strategy saves checkpoints every $\sqrt{n}$ steps, where n is number of layers (depth) of the DNN \citep{GC2019, GcMediumArticle}. This strategy ensures that the computation time still scales linearly to the depth of the DNN, while reducing GPU memory requirements from linear to square root of the depth of the DNN. Thus, Gradient Checkpointing allows to fine-tune much larger LLMs than what is possible with an additional 20\%-30\% increase in fine-tuning time \citep{GcMediumArticle}. 

\subsection{Low-Rank Adaptation}
Low-Rank Adaptation (LoRA) reduces the number of trainable parameters which in turn lowers GPU memory requirements during fine-tuning. It achieves this by freezing the pre-trained model weights and subsequently injecting trainable rank decomposition matrices into each of the chosen layers of the Transformer architecture \citep{hu2021lora}. The number of trainable parameters are reduced by orders of magnitude, bringing down the cost of fine-tuning, while also preserving the quality of the results. As an example, with LoRA fine-tuning using rank set to 64, the number of trainable parameters for a 70 billion parameter model is reduced to about 131 million parameters ($\sim$0.19\% of the original model size).

\subsection{Zero Redundancy Optimizer}
DeepSpeed’s Zero Redundancy Optimizer (ZeRO) \citep{rajbhandari2020zero} is a memory optimization technique that provides the benefits of model and data parallelism, while alleviating the limitations of both. ZeRO-powered data parallelism (ZeRO-DP) partitions the model states – parameters, gradients, and optimizer states – across the data-parallel processes and uses a dynamic communication schedule to share the necessary model states across processes. ZeRO-DP provides three optimization stages, which successively provide larger and larger memory reduction while incurring some runtime overhead. ZeRO-DP based fine-tuning can achieve memory reduction from 4 to 8 times for Stages 1 and 2 respectively, and all the way up to linear memory reduction for Stage 3 (given a large number of GPUs). This does come at the expense of runtime, especially for ZeRO-DP Stage 3. ZeRO optimization is further enhanced by the inclusion of ZeRO-Offload and ZeRO-Infinity \citep{ren2021zerooffload, rajbhandari2021zeroinfinity}, which enable offloading the optimizer states and model parameters to CPU respectively. While ZeRO-Offload is available across all three ZeRO-DP stages, ZeRO-Infinity is available only for stage 3. Without using DeepSpeed's ZeRO kind of optimizations, many large models (several billions of parameters) are practically impossible to train using just a handful of V100 or A100 GPUs.

\subsection{FlashAttention}
FlashAttention helps achieve attention calculations in linear instead of quadratic complexity with respect to the sequence/context length \citep{dao2022flashattention}. It leverages tiling and recomputation techniques to significantly speed up attention computation. It makes judicious use of the Static Random Access Memory (SRAM), which is the most expensive but smallest sized memory unit within the GPU, minimizing reads and writes between SRAM and HBM. In FlashAttention-2 (FA2), there are further optimizations to reduce the slower non-matrix multiplication operations and to parallelize forward and backward passes along the sequence length dimension (in addition to the batch and number of heads dimensions) \citep{dao2023flashattention2}.  

One key feature of all these optimizations is that they are orthogonal to each other. Therefore, they can be combined to get incremental memory and runtime benefits for the users.

\section{Theoretical Analysis of GPU Memory Requirements}\label{tagmr}
\begin{figure}[!ht]
\centering
\includegraphics[scale=0.8]{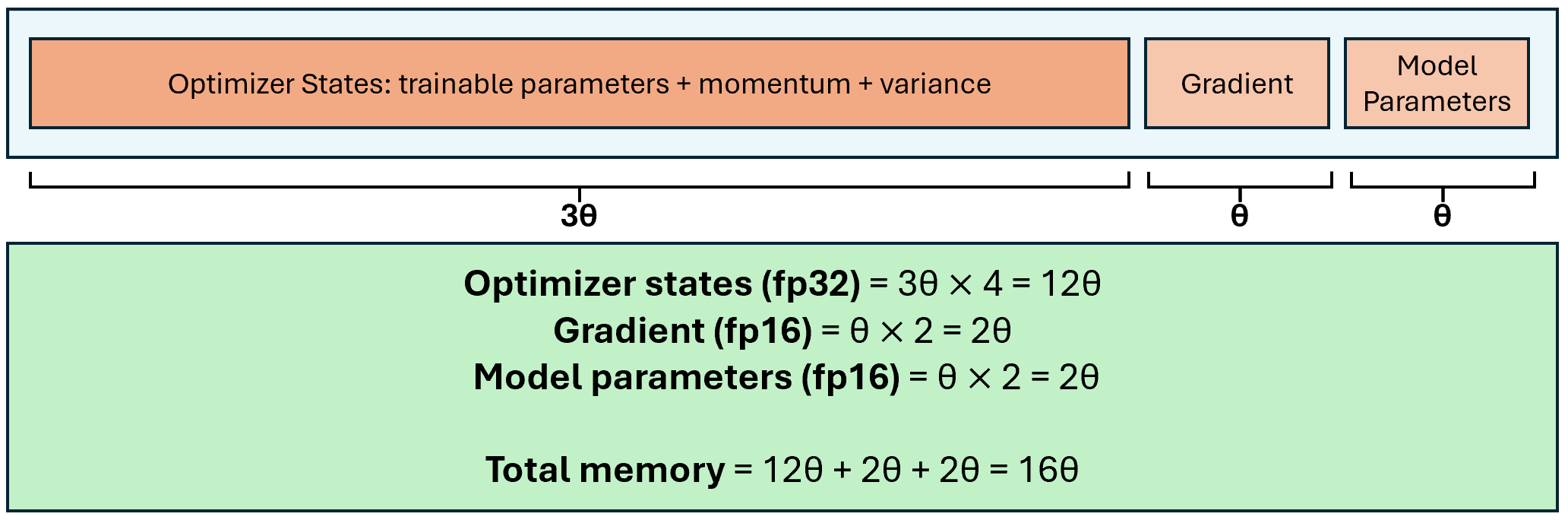}
\caption{\label{fig:Model_State_Mem} Model state memory for a model with $\theta$ parameters, when fine-tuned using Adam optimizer under mixed-precision setting. Model state comprises of optimizer state, gradients and model parameters. Total model state memory, with no optimization enabled, adds upto 16$\theta$ bytes.}
\label{MSM}
\end{figure}
In this section, our goal is to help the reader gain a fundamental understanding of the calculations influencing GPU memory consumption during the training process. Following three variables primarily affect GPU memory during training:
\begin{enumerate}[A.]
\item Model states – including model parameters, gradients, and optimizer states
\item Activations – which represent the intermediate computational results
\item Temporary buffers and fragmentation
\end{enumerate}
\textbf{Model states memory}: We use the memory computations for model states in \citep{rajbhandari2020zero} to estimate (A). Here we briefly discuss the model state memory computation shown in Figure \ref{MSM}. During mixed-precision training with Adam optimizer, trainable parameters, momentum and variance are maintained in full-precision (32 bits). Assuming $\theta$ parameters, the memory adds up to $(\theta$ + $\theta$ + $\theta) \times 4$ = 12$\theta$ bytes. Thus, K=12 is the Adam optimizer specific constant multiplier term used in memory calculations. Further, the parameters and gradients are maintained in half-precision during forward and backward passes which incur additional memory requirement of 2$\theta$ + 2$\theta$ = 4$\theta$ bytes. ZeRO-DP stage determines which model states (parameters, gradients, optimizer states) are partitioned across GPUs leading to memory savings.

\textbf{Activation memory}: Activation memory per transformer layer, in the absence of model parallelism,  can be estimated using Equation 1 in Section 4 of the paper \citep{korthikanti2022reducing}. We rely on the same calculations to determine an approximate upper bound on the activation memory requirements for (B). Activation memory per layer can be computed as:
$$ s\times b \times h \times \left (34 + \left (5 \times \frac{a \times s}{h} \right ) \right ) $$
Here, $s$ is the sequence length, $b$ is the microbatch size, $h$ is the hidden dimension, $a$ is the number of attention heads. The total activation memory is then computed as Number of transformer layers $\times$ Activation memory per layer.
\begin{table}[!ht]
\begin{center}    
\begin{tabular}{ |p{1.5cm}|p{2.5cm}|p{2.5cm}|p{2.5cm}|p{2.5cm}|p{2.5cm}|  }
 \hline
ZeRO-DP Stage & Model State memory & Activation memory & Theoretical GPU memory (no Offload) & Empirical GPU memory (no Offload) & Empirical GPU memory (with Offload) \\[0.5ex] 
 \hline
 ZeRO-1 & 4 $\times$ 7 + (12/8) $\times$ 7 & 1.48 & 39.98 & 37.15 & 28.38 \\ 
 \hline
 ZeRO-2 & 2 $\times$ 7 + (14/8) $\times$ 7 & 1.48 & 27.73 & 30.52 & 17.43 \\
 \hline
 ZeRO-3 & (16/8) $\times$ 7 & 1.48 & 15.48 & 18.65 & 4.97 \\ [1ex] 
 \hline
\end{tabular}
\end{center}
\caption{Comparison between theoretical and empirical GPU memory allocated (in GB) across the three stages of ZeRO-DP. Fine-tuned model is Llama 2 7B using 8xA100 GPUs (each with 80 GB of HBM). The theoretical estimates can be used to approximate the empirical memory requirements before running the fine-tuning jobs.}
\label{ThEMComp}
\end{table} 

Table \ref{ThEMComp} shows our results of comparing the theoretically expected and empirical GPU memory allocated for fine-tuning Llama 2 7B across ZeRO-DP stages. We use the notation ZeRO-1, ZeRO-2 and ZeRO-3 to denote ZeRO-DP stages 1, 2 and 3  respectively. The theoretical memory is computed by summing the memory from model states and activations. We use the following model default parameters: number of transformer layers=32, hidden dimension size $h$=4096 and attention heads $a$=32 to theoretically compute the total activation memory of 1.48 GB. The empirical GPU memory refers to the peak GPU memory allocated during fine-tuning. We show empirical memory with and without CPU offloading of optimizer states. The experimental setup details are described in subsection \ref{expsetup}. 

Our results show that the theoretical calculations can serve as a rough estimate for GPU memory consumption (without CPU off-loading) during the fine-tuning process. Calculating this estimate prior to running fine-tuning jobs can help predict the memory requirements and efficiently plan resource allocation. We note that CPU offloading of optimizer states can scale down the memory usage upto 4 times than when off-loading is not enabled. Thus, CPU offloading can allow efficient utilization of both GPU and CPU resources for fine-tuning multi-billion parameter models which is a memory-intensive process.   

Note that devoid of any ZeRO-DP optimization, full fine-tuning of a 7 billion parameter model using conventional data parallelism would require more than 112 (16 $\times$ 7) GB of GPU memory. Such memory requirement far exceeds the 80 GB of GPU HBM available with A100s.

\section{Experiments}\label{exps}
We begin this section by analyzing the impact of fine-tuning optimizations on memory and runtime. The results of our analysis are used to recommend a set of optimizations that can act as a balanced default to optimize for memory and runtime during fine-tuning. Subsequently, we delve into fine-tuning large models (with tens of billions of parameters) and explore the optimizations which make such fine-tuning possible. Following this, we examine the role of FlashAttention-2 during fine-tuning of LLMs on long context data. Lastly, we investigate the specific case of fine-tuning with resource-constrained GPUs, specifically V100s. Our goal is to identify the fine-tuning optimizations that can enable efficient fine-tuning across various model sizes and context lengths, even with limited resources.

\subsection{Setup}\label{expsetup}
We fine-tune models from Llama 2 (7B, 13B, 70B) \citep{touvron2023llama} and Falcon (180B) \citep{almazrouei2023falcon} families on Causal Language Modeling task. Following are the details of our experimental setup:
\begin{itemize}
\item \textbf{Data}: Sampled Samsum dataset \citep{Gliwa_2019} is used for fine-tuning in all the experiments.
\item \textbf{Compute}: Standard\_ND40rs\_v2 (8xV100) \citep{nd40rsv2} and Standard\_ND96amsr\_A100\_v4 (8xA100) \citep{nda100_80gb} are used as the GPU computes.
\item \textbf{Optimizer}: All our experiments use mixed-precision setting and AdamW optimizer ($\beta_1$=0.9 and $\beta_2$=0.99) with a linear scheduler and learning rate of 4e-4.
\item \textbf{Sequence Length, Batch Size and Epochs}:  The sequence length is typically set to 256. In studies where we explore the impact of sequence length, length of input sequences is extended through padding. The effective batch size used in all our experiments is 8. Note that gradient accumulation can be used to increase the effective batch size, while leading to efficient use of available GPU memory. All models are fine-tuned for a single epoch.
\item \textbf{LoRA}: In experiments where LoRA is used, LoRA rank is set to 64 and alpha is set to 32.
\item \textbf{CPU Offload}: Unless explicitly noted, we use ZeRO-Offload to offload optimizer states and computations to CPU during fine-tuning. This helps train multi-billion parameter models using computational and memory resources of available GPUs and CPUs.
\end{itemize}

In our experiments, GPU memory usage is measured by the peak GPU memory allocated across all GPUs used in the fine-tuning process.

\subsection{Best Default Optimizations to Balance Memory and Runtime}
\begin{figure}[ht!]
    \centering
     \includegraphics[width=\textwidth]{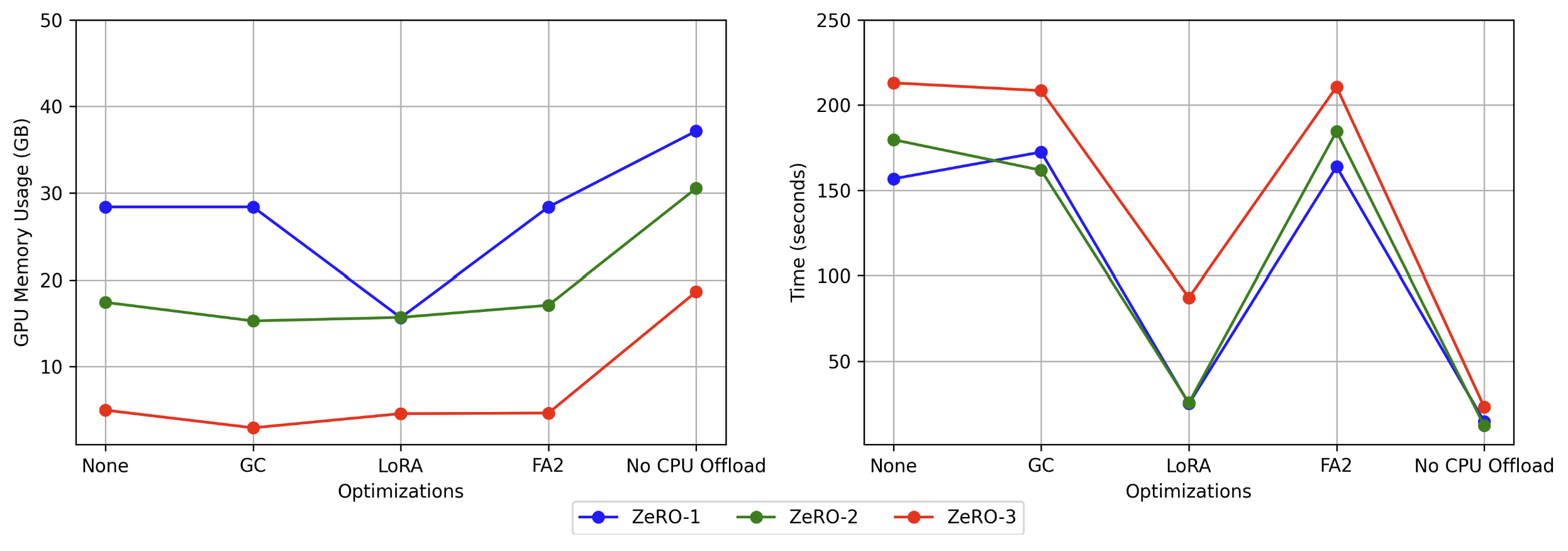}
     \caption{GPU memory usage and fine-tuning runtime for different optimization configurations across ZeRO-1, ZeRO-2 and ZeRO-3 for Llama 2 7B. Using LoRA with ZeRO-2 provides the best balance between memory usage and runtime.}
     \label{fig:best_defaults}
\end{figure}

We have already established that for a relatively small 7B parameter model, it is infeasible to fine-tune without any optimizations, even when using A100s. Therefore, we use DeepSpeed's ZeRO which provides both model and data parallelism. We examine the set of fine-tuning optimizations from section \ref{oot} that should be enabled by default in a fine-tuning framework. Our criteria for selecting such optimizations is based on their ability to strike an optimal balance between memory usage and runtime in conjunction with ZeRO-DP. 

We fine-tune Llama 2 7B on one node (8xA100 GPUs) with five distinct configurations: (a) no optimizations among GC, LoRA and FA2 enabled (b) with GC (c) with LoRA (d) with FA2 and (e) without CPU offloading. Each configuration is enabled with all three ZeRO-DP stages. We chose A100 GPUs for these  experiments due to their substantial memory capacity of 80 GB. This ample memory allows us the flexibility to experiment with various optimizations by toggling them on and off, while avoiding out-of-memory failures. 

Figure \ref{fig:best_defaults} shows the GPU memory usage and runtime across each configuration and ZeRO-DP stage. Based on the results, we draw the following conclusions:
\begin{enumerate}
\item The overall best combination of optimizations from the standpoint of both runtime and memory is configuration (c) i.e. ZeRO-DP + LoRA. Since LoRA reduces the total trainable parameters, it saves memory and cuts down the run-time.

Within ZeRO-DP + LoRA configuration, memory usage as well as runtime are approximately similar between ZeRO-1 and ZeRO-2 stages. When using LoRA (with rank=64), trainable parameters are reduced to roughly 33.6 million for the 7B model. The optimizer states and gradients of the trainable parameters occupy a minimal portion of the GPU memory. The memory consumption in ZeRO-1 and ZeRO-2 is dominated by the total model parameters (7 billion) that account for 14 GB of memory in half-precision.
\item Configurations (a) and (d) perform expectedly across ZeRO-DP stages i.e. the memory consumption drops and runtime increases from stage 1 $\rightarrow$ 2 $\rightarrow$ 3. For configurations (b) and (e), ZeRO-1 runtime is marginally higher than ZeRO-2, and the memory usage follows the expected trend of reducing with increase in ZeRO-DP stage.
\item Disabling CPU offloading in (e) provides the fastest runtime at the expense of requiring 2-4 times more GPU memory.
\end{enumerate}

We conclude that the combination of ZeRO-2 and LoRA serves as an excellent default, given its ability to maintain a balance between memory usage and runtime. Furthermore, ZeRO-2 can handle larger models than ZeRO-1, which in turn helps support a broader range of model sizes.  For certain use-cases such as complex question-answering tasks or building chatbots, the user may prefer full fine-tuning of the model. Thus, the choice to use LoRA is at the discretion of the user. Therefore, we recommend to activate ZeRO-2 by default during the fine-tuning process.

\subsection{Fine-tuning Large Models}
Fine-tuning large models such as Llama 2 70B or Falcon 180B can often run into out-of-memory (OOM) errors if the right set of optimizations are not enabled. Enabling ZeRO-3 is a viable alternative in such scenarios though it comes at the expense of increased runtime. Let's examine the model state memory term in ZeRO-3, fine-tuned over $M$ nodes with $N$ GPUs per node:
$$\text{Model states memory} = \frac{2\theta + 2\theta + 12\theta}{M \times N} = \frac{16\theta}{M \times N} \times 10^9 \text{ (bytes)} = \frac{16\theta}{M \times N} \text{ (GB)}$$ 
Here, $\theta$ denotes the total trainable parameters (in billions). Lets assume a standard configuration of $N$=8 GPUs per node. This makes the model states memory $\frac{2\theta}{M}$. This means scaling the number of nodes can effectively neutralize the increase in model size. For example, full fine-tuning of $\theta$=100B parameter model over $M$=5 nodes occupies 40 GB of model state memory which can fit within A100s with 80 GB HBM. In practice, large models with hundreds of billions of parameters are often fine-tuned with LoRA, and using CPU off-loading with ZeRO-3. This brings down the memory requirements further to a manageable scale.

\begin{table}[!ht]
\begin{center}
\begin{tabular}{|c|c|c|}
\hline
Llama 2 Model & GPU Memory Usage (GB) & Time (seconds)\\\hline
70B & 15.54 & 278.26 \\
\hline
13B & 6.22 & 75.67 \\
\hline
7B & 4.74 & 55.18 \\
\hline
\end{tabular}
\end{center}
\caption{\label{tab:model-variants}GPU memory usage and fine-tuning time for Llama 2 Models with ZeRO-3 + LoRA on 8xV100 GPUs.}
\end{table}

We were able to fine-tune Llama 2 70B using a combination of ZeRO-3 + LoRA on a single node with 8xV100 GPUs (32 GB HBM). Table \ref{tab:model-variants} shows the actual GPU memory consumed to be around 15.54 GB for the 70B model in our experiments. Additionally, it can be seen that the GPU memory is under-utilized for smaller models like Llama 2 7B and 13B with ZeRO-3 + LoRA. Consequently, our study shows that for fine-tuning larger models with tens or hundreds of billions of parameters, enabling ZeRO-3 is indispensable. Furthermore, when used in conjunction with LoRA, the overall memory requirements significantly reduce in practice. 

For smaller models, ZeRO-3 + LoRA is evidently excessive as it can lead to under-utilization of GPUs along with increased runtime. We refer the reader to subsection \ref{ftgpuconst} where we explore the appropriate optimization choices for different model sizes. 

\subsection{Long Context Fine-tuning}
\begin{figure}[ht!]
    \centering
     \includegraphics[width=\textwidth]{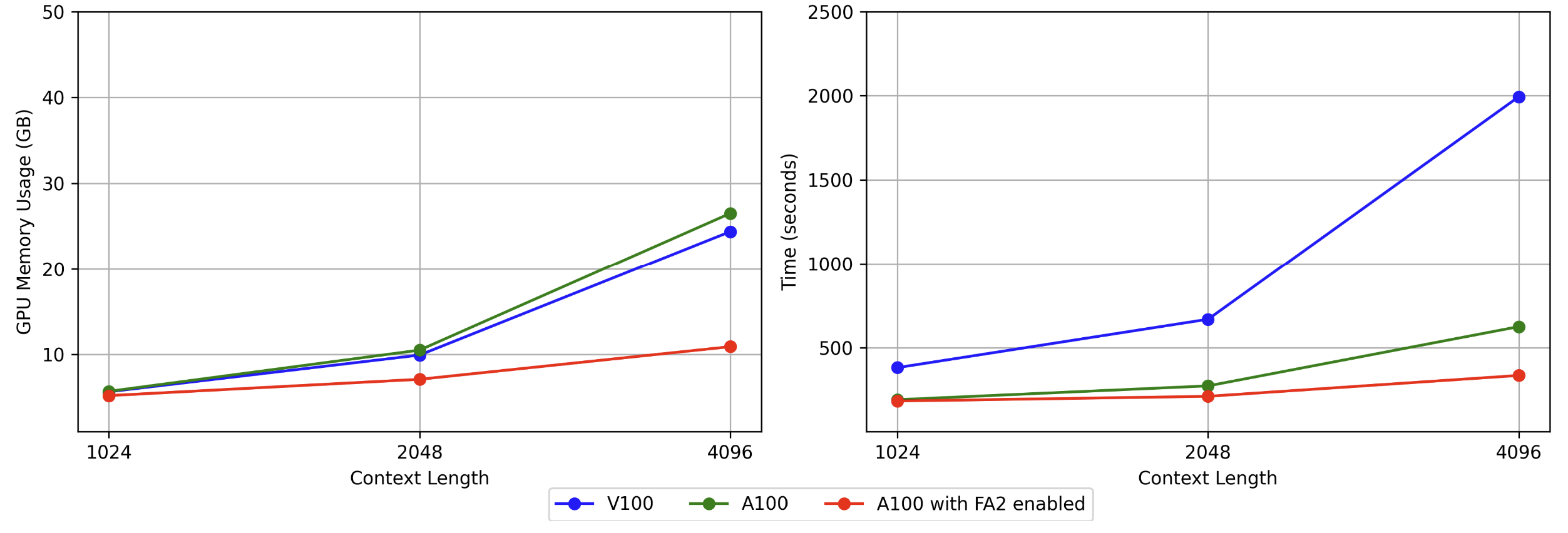}
     \caption{Impact of varying context length on GPU memory usage and fine-tuning time with and without FlashAttention-2 for Llama 2 70B. Enabling FlashAttention-2 on A100s significantly lowers the memory consumption and runtime for larger context lengths such as 4096.}
     \label{fig:long_context}
\end{figure}

Context length is a critical aspect when fine-tuning LLMs. This is especially true when fine-tuning with datasets comprising of very long sequences of text. Without FlashAttention-2 (FA2), attention computations scale quadratically in sequence length. Since V100 GPUs do not support FA2, the impact of using longer context lengths can be significant on GPU memory consumption. To study the impact of long context lengths during fine-tuning process, we plot memory and runtime on A100s (with and without FA2 enabled) and V100s (without FA2) across different context lengths. Specifically, we fine-tune Llama 2 70B model with context-lengths of 1024, 2048 and 4096 separately on 8xV100 and 8xA100 GPUs. LoRA and Gradient Checkpointing are enabled on top of ZeRO-3 during fine-tuning.

Figure \ref{fig:long_context} demonstrates the impact of enabling FA2 on GPU memory and runtime. Our results show an expected trend as context length is increased to 4096. There is a significant reduction in GPU memory usage and runtime when using FA2 than without it. As seen in Figure \ref{fig:long_context}, this trend is consistent when comparing runs with and without FA2 on two A100 GPUs, or on A100 and V100 GPUs respectively. Our conclusion is as follows. High-performance GPUs like A100s, due to their larger HBMs, can support relatively long context fine-tuning without FlashAttention-2. However, it is optimal to activate FlashAttention-2 on compatible GPU architectures as that allows additional memory savings and reduced runtime. This becomes particularly important when training on long text data with models that allow fine-tuning with larger context lengths.

\subsection{Fine-tuning under GPU Resource Constraints}\label{ftgpuconst}
\begin{figure}[ht!]
\centering
\includegraphics[scale=0.7]{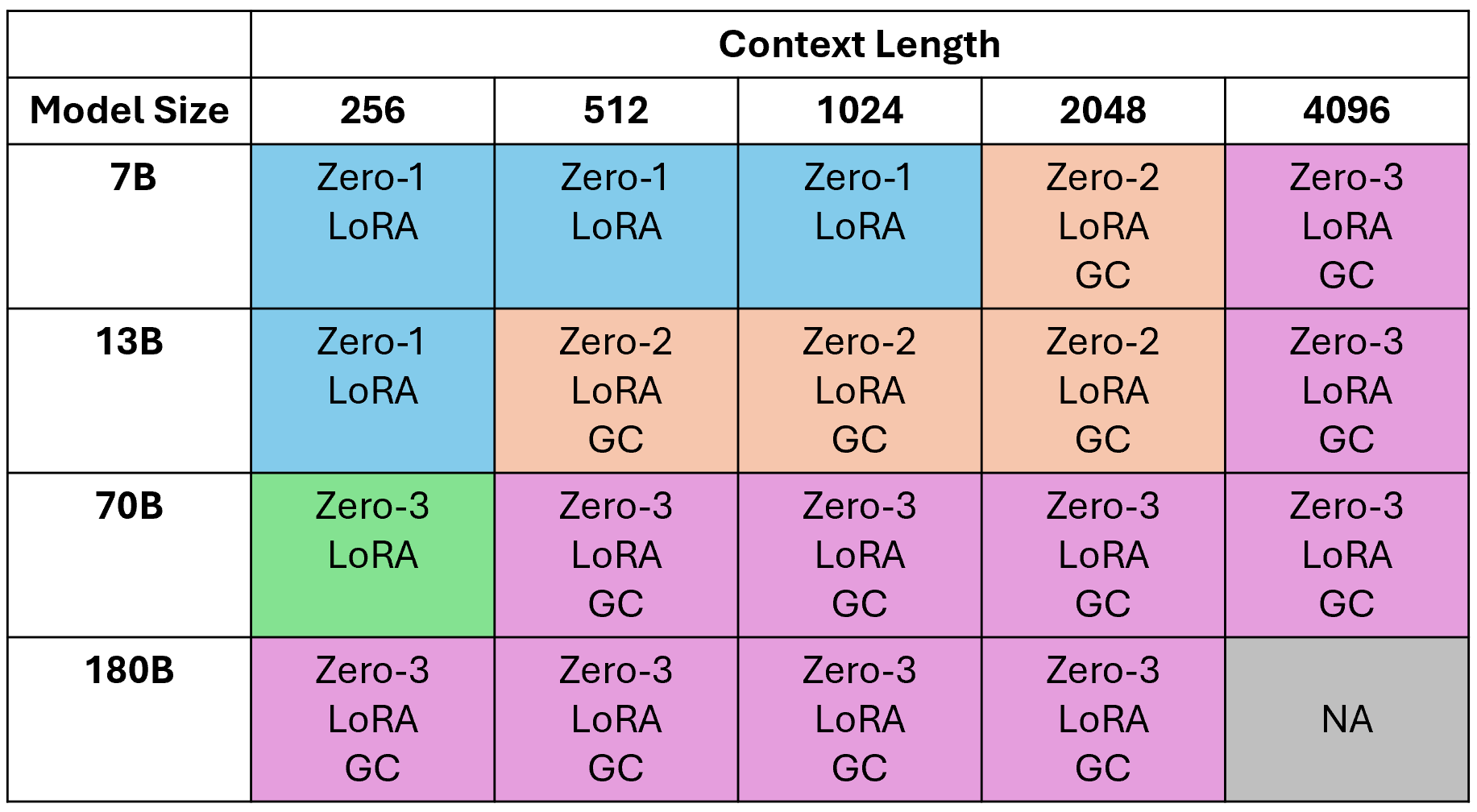}
\caption{\label{fig:combo2}Optimal configurations for fine-tuning LLMs of different sizes when using V100 GPUs. All Llama 2 experiments were run using 8xV100s, whereas Falcon 180B required 16xV100s. FlashAttention-2 is omitted as it is not supported on V100s.}
\end{figure}
In this section, we discuss fine-tuning LLMs (between 7B to 180B parameters) with limited resources. We define the resource constraint as follows:
\begin{itemize}
\item Low resource GPUs are available with limited HBM.
\item Number of GPUs available is small.
\end{itemize}
We fine-tune Llama 2 (7B, 13B, 70B) and Falcon (180B) models over five different context lengths: [256, 512, 1024, 2048, 4096]. Adhering to the resource constraint specifications, our experiments are done on a Standard\_ND40rs\_v2 (8xV100 GPUs) with 32 GB of HBM. The only deviation from this setup is for Falcon 180B model. Due to its large size, we use two nodes (16xV100 GPUs) during fine-tuning. Figure \ref{fig:combo2} illustrates the preferred set of optimizations that should be enabled to meet GPU memory requirements while optimizing for runtime. Note that while there can be other combinations that allow successful fine-tuning, we select those which minimize the fine-tuning time. As an example, users can choose to enable all optimizations, but often they are not all going to be warranted, and may unnecessarily increase the fine-tuning time.

Key insights from our results are:
\begin{enumerate}
\item Combining model and data parallelism is a prerequisite even for fitting models as small as 7B parameters on a limited number of GPUs, unless we leverage quantization.
\item As context length or model size increases, it becomes important to shift to a higher ZeRO-DP stage to facilitate fine-tuning. This trend is clearly visible in Figure \ref{fig:combo2}, where a move from left to right or top to bottom corresponds to a higher ZeRO-DP stage. 
\item Gradient Checkpointing (GC) is an effective memory-saving optimization technique, especially for large models. As depicted in Figure \ref{fig:combo2}, enabling GC for Llama 2 13B and 70B allowed support of context lengths from 512 to 4096, which would have not been possible otherwise. In a similar vein, GC is essential for enabling fine-tuning of Falcon 180B on all our examined context lengths.
\end{enumerate}

Although FlashAttention-2 was not supported on V100 GPUs, our experimental results suggest that it should always be enabled on supported architectures. It is worth mentioning that the configurations outlined in Figure \ref{fig:combo2} can be applied for fine-tuning on GPUs with higher HBMs than V100s. While these configurations may not be optimal for all GPUs, our findings can be extended to other GPUs with further investigation.

\section{Conclusion and Future Work}\label{cfw}
In this paper, we analyzed state-of-the-art fine-tuning optimizations and their impact on memory and runtime during fine-tuning. Our results show that ZeRO-2 + LoRA is a reliable optimization default that typically provides the best balance between memory usage and fine-tuning runtime. Using the right set of optimizations, such as ZeRO-3 + LoRA + GC, we successfully fine-tuned models as large as Falcon 180B. Recognizing the need for fine-tuning LLMs under GPU constrained settings, we introduced an optimization matrix across model-size $\times$ context length. The optimization matrix can guide the user to select the right optimization mixture that enables fine-tuning for a given use case. We conclude that optimizations such as DeepSpeed's ZeRO are essential for fine-tuning multi-billion parameter models as they support data and model parallelism, along with CPU off-loading. In conjunction with ZeRO, it becomes necessary to select additional optimizations to avoid out-of-memory failures while balancing fine-tuning runtime.

Following directions can be considered for future work. 1) Incorporate 4-bit and 8-bit Quantization analysis \citep{dettmers2023qlora} 2) Explore optimal configurations for fine-tuning Small Language Models (SLMs) such as Phi-2 \citep{phi2blog} and Phi-3-mini \citep{abdin2024phi3} and 3) Examine fine-tuning strategies with larger context length of up to 128K.

\section*{Acknowledgements}
We thank the Azure Machine Learning Fine-tuning team for supporting this work. We thank Omkar More and Bala Venkataraman for helping review the work.

\bibliographystyle{plainnat}
\bibliography{ft_optimizations_study}

\end{document}